\pdfoutput=1
\documentclass[11pt]{article}

\usepackage{acl}
\usepackage{times}
\usepackage{latexsym}
\usepackage{amsmath}
\usepackage[T1]{fontenc}
\usepackage[utf8]{inputenc}
\usepackage{booktabs}
\usepackage{microtype}
\usepackage{inconsolata}
\usepackage{graphicx}
\usepackage{dashrule}
\usepackage{soul, color, xcolor}

\begin{document}

\title{GW-MoE: Resolving Uncertainty in MoE Router\\ with Global Workspace Theory}

\author{
\small{
$^{1}$Haoze Wu\thanks{Equal contribution}, 
$^{2,3}$Zihan Qiu$^*$\thanks{Work done while interning at INF Technology.},  
$^{3}$Zili Wang, 
$^{2}$Hang Zhao\thanks{Corresponding authors}, 
$^{4}$Jie Fu$^{\text{\ddag}}$ 
} \\
\small{$^{1}$Zhejiang University $\,\,\,$ $^{2}$Tsinghua University $\,\,\,$
$^{3}$INF Technology $\,\,\,$
$^{4}$Hong Kong University of Science and Technology}\\
\small{
\texttt{waithz@zju.edu.cn, qzh11628@gmail.com, ziliwang.do.gmail.com,}}\\
\small{
\texttt{hangzhao@tsinghua.edu.cn, jiefu@ust.hk}}
}

\maketitle

\begin{abstract}
    Mixture-of-Experts (MoE) has been demonstrated as an efficient method to scale up models.
    By dynamically and sparsely selecting activated experts, MoE can effectively reduce computational costs.
    Despite the success, we observe that many tokens in the MoE models have uncertain routing results. 
    These tokens have nearly equal scores for choosing each expert, and we demonstrate that this uncertainty can lead to incorrect selections. 
    Inspired by the Global Workspace Theory (GWT), we propose a new fine-tuning method, GW-MoE, to address this issue. 
    The core idea is to broadcast the uncertain tokens across experts during fine-tuning.
    Therefore, these tokens can acquire the necessary knowledge from any expert during inference and become less sensitive to the choice. 
    GW-MoE does not introduce additional inference overhead.
    We validate that GW can mitigate the uncertain problem and consistently improve in different tasks (text classification, question answering, summarization, code generation, and mathematical problem solving) and model sizes ($650$M and $8$B parameters).
    Our code is publicly available at \url{https://github.com/WaitHZ/GW-MoE}.
\end{abstract}

\section{Introduction}
\label{sec:intro}

In recent years, large language models (LLMs) have developed rapidly~\citep{devlin2019bert, touvron2023llama, openai2024gpt4} and have been widely applied in numerous fields, including education, healthcare, and smart transportation~\cite{dan2023educhat, li2023chatdoctor, zheng2023chatgpt}. 
The impressive capabilities of LLMs can be mainly attributed to the increased model scale. 
However, continuously increasing the scale of LLMs raises the difficulty of model deployment and poses challenges for promoting LLMs within the open-source community. 
As a result, the sparse activation moel MoE has been receiving increasing attention~\cite{shazeer2017outrageously, fedus2022switch}.

\begin{table}
  \centering
  \scalebox{0.85}{
  \begin{tabular}{c|ccc}
    \toprule
        & \textbf{Bottom}$\boldsymbol{5\%}$ & \textbf{Average} &  \; \textbf{Top}$\boldsymbol{5\%}$ \; \\
    \midrule
    Mixtral-$8\times 7$B & $0.63$ & $0.79$ & $\boldsymbol{0.93}$ \\
    DeepSeek-$16$B & $0.81$ & $0.88$ & $\boldsymbol{0.93}$ \\
    JetMoE-$8$B & $0.88$ & $\boldsymbol{0.94}$ & $\boldsymbol{0.97}$ \\
    \bottomrule
    
  \end{tabular}
  }
  \caption{
  \textbf{Uncertain tokens are not uncommon in MoE models.}
  We calculate the expert selection entropy from routing scores in the first layer of three common MoE models.
  The entropy is normalized in  
  $[0, 1]$ follow Sec~\ref{sec:entropy-uncertain}, with value $1$ corresponding to a uniform score distribution. 
  Some tokens have almost uniform expert selection. We call them `uncertain tokens'.}
  \label{tab:common-moe}

  \vskip  -0.22in
\end{table}

MoE models reduce computational costs by sparsely activating only a small number of model parameters for a single input.
In existing works~\cite{fedus2022switch, dai2024deepseekmoe, shen2024jetmoe}, the proportion of activated experts is typically $\frac{1}{8}$ or $\frac{1}{4}$.
Specifically, the router (usually a linear layer) in MoE models outputs the score of selecting each expert based on the input, and those with the highest scores will be selected. 
However, when testing some common open-source MoE models with billion  parameters~\cite{jiang2024mixtral, dai2024deepseekmoe, shen2024jetmoe} on Alpaca~\cite{alpaca}, we notice that router assign some tokens to experts with almost uniform scores.
We use normalized entropy to measure the uncertainty of the expert selection of tokens.
Normalized entropy is calculated by summing the products of each outcome's probability and its logarithm, divided by the logarithm of the number of outcomes.
When the value approaches $1$, it indicates that the score distribution output by the router is close to be uniform.
As shown in Tab~\ref{tab:common-moe}, a subset of tokens has a normalized entropy that is greater than $0.9$ in all three MoE models.
We use \textbf{uncertain tokens} for such phenomena in the rest of the passage.

\begin{figure}[htbp]
    \centering
    \includegraphics[width=0.48\textwidth]{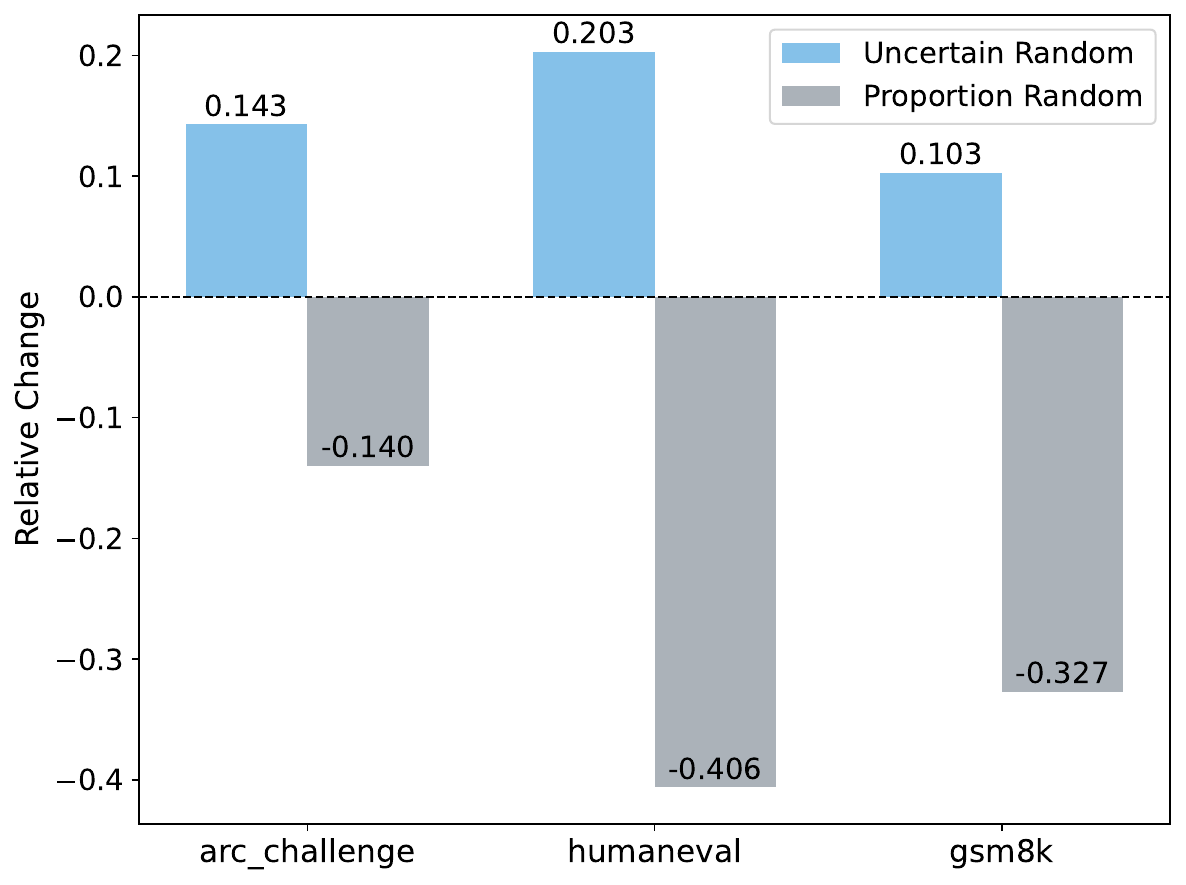}
    \caption{\textbf{Randomly selecting experts for uncertain tokens can give better results}. 
    We let the uncertain tokens (entropy greater than $2.0$) in the last layer of JetMoE randomly select experts, and the average results (blue) from multiple experiments on three tasks are better than those obtained by using the Top-$K$ operator to select experts (dashed line). 
    To further verify, we let the same proportion of arbitrary tokens randomly select experts and observe that the results (gray) are worse than uncertain random. The metrics for each task are the same as those in Sec~\ref{sec:IT}.}
    \label{fig:random}

    \vskip  -0.1in
\end{figure}

Additionally, we demonstrate on JetMoE-8B that randomly selecting experts for uncertain tokens can outperform the choices made by the Top-$K$ operator, as shown in Fig~\ref{fig:random}.
More unfortunately, vanilla fine-tuning increases the number of uncertain tokens. 
We find that $72\%$ of the tokens in the JetMoE-8B remain uncertain after fine-tuning, and the number of uncertain tokens is $3.4$ times larger than before.

Why is it important for tokens to select the correct expert? 
~\citet{geva2021transformer, qiu2024empirical} suggest that the FFN layer, commonly replaced by MoE in transformers, acts as a key-value memory network.
In the MoE models, each expert acts as an independent memory block, and the router determines which one to access for each token.
If the router is uncertain for some tokens, these tokens may fail to access the necessary knowledge.

To solve this problem, we take inspiration from human brains. 
GWT~\citep{GWT} suggests that there are independent functional modules in the human brain for processing different neural signals. 
For complex signals, modules can cooperate by broadcasting information through a global workspace.
When learning new knowledge, this broadcasting mechanism helps form long-term memory and strengthens the stability and accessibility of knowledge recall.
Similar to the human brain, each expert in MoE models can be seen as a functional module; like complex signals, uncertain tokens also need to be more easily accessible.
We believe that GWT provides valuable insights for fine-tuning MoE models. 

\begin{figure*}[htbp]
    \vskip  -0.3in

    \centering
    \includegraphics[width=0.28\textwidth]{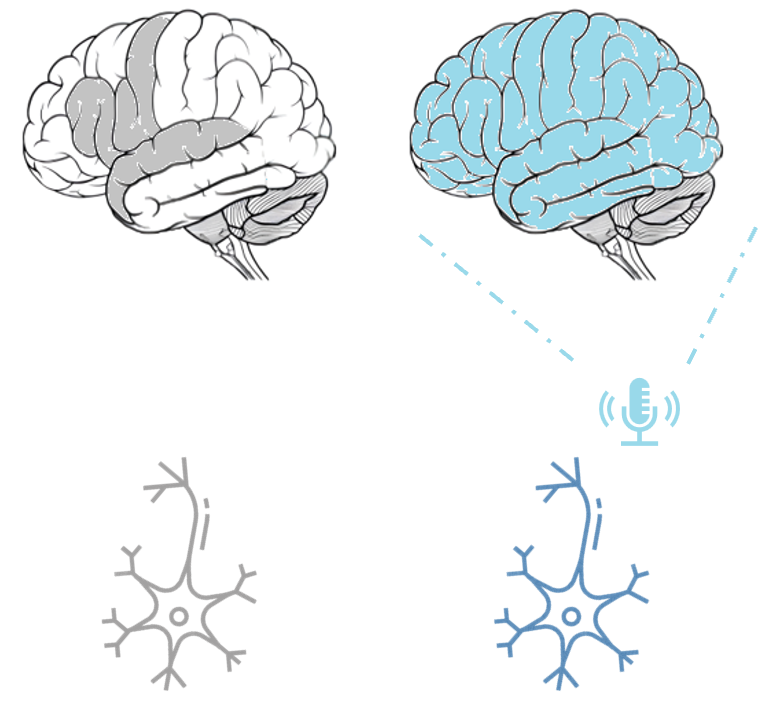}
    \hspace{0.4in}
    \includegraphics[width=0.55\textwidth]{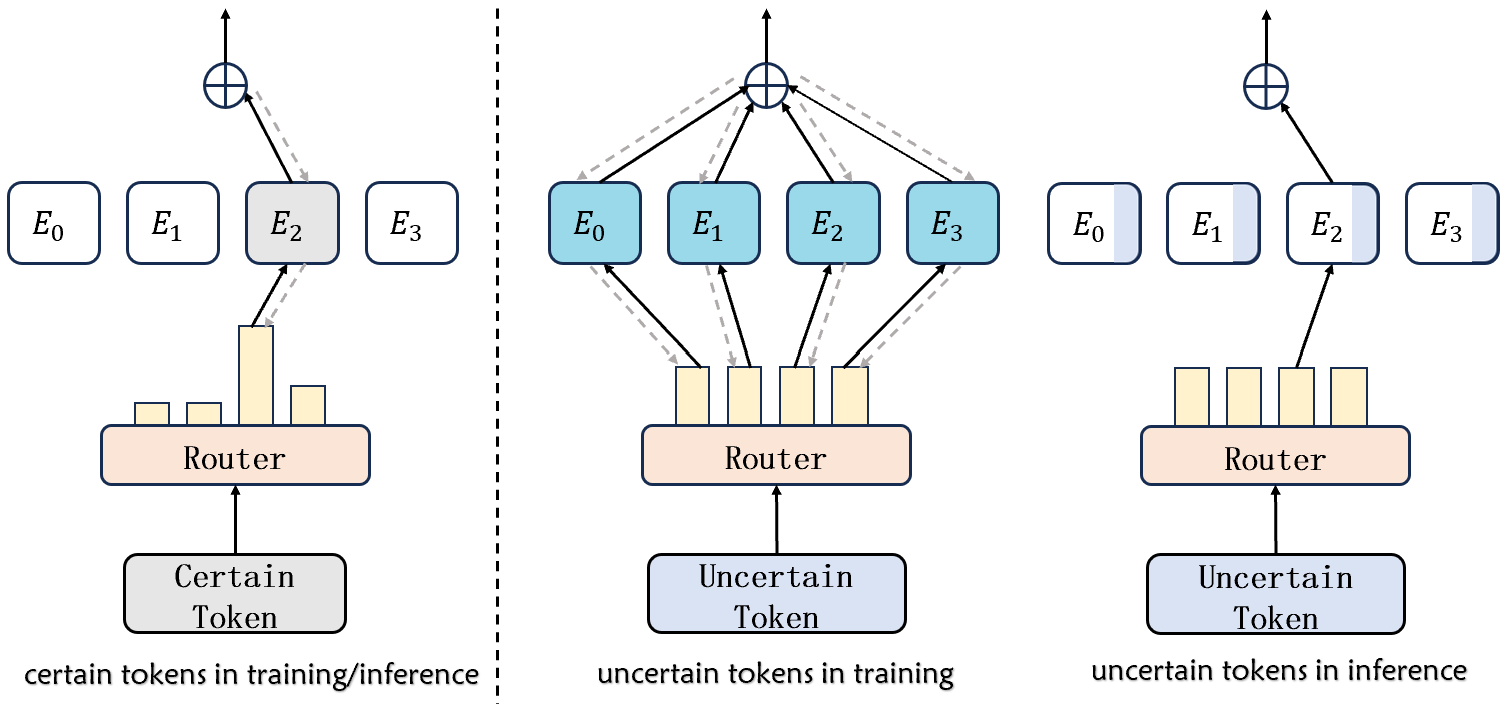}
    \caption{Overview of GW-MoE. \textbf{Left:} Based on the GWT, some neural signals (grey) only need to activate a single functional module in the human brain, while others (blue) will use the global workspace to broadcast information, facilitating cooperation between modules. \textbf{Right:} GW-MoE is inspired by GWT. When the router's output score is nearly uniform, those tokens (blue) are called uncertain tokens and are broadcast to all experts during fine-tuning; during inference, since all experts have learned the knowledge of uncertain tokens, these tokens can obtain the necessary information from any expert. The rest (grey) are certain tokens, routed to the Top$K$ experts during both inference and fine-tuning, following standard MoE.}
    \label{fig:overview}

    \vskip  -0.15in
\end{figure*}

Based on this, we propose a novel method for fine-tuning MoE models called \textbf{G}lobal \textbf{W}orkspace tuning for \textbf{M}ixture-of-\textbf{E}xperts (GW-MoE). 
During fine-tuning, we broadcast uncertain tokens to all experts, allowing each to learn the relevant knowledge, so that during inference, uncertain ones can obtain the necessary knowledge from any expert, as shown in Fig~\ref{fig:overview}.
Because all experts have learned knowledge of uncertain tokens during fine-tuning, GW-MoE does not introduce any additional inference overhead.
This ensures that the model remains efficient in various applications.

 We evaluate GW-MoE across different model scales (from hundreds of millions to several billions parameters) on various tasks, including natural language understanding, question answering, summarization, mathematical problem-solving, and code generation. 
 Extensive experimental results have demonstrated that our method consistently outperforms standard fine-tuning.

 Summarizing, our core contributions are:

 \begin{itemize}
    \setlength{\itemsep}{0.5pt}
    \setlength{\parsep}{0.5pt}
    \setlength{\parskip}{0.5pt}
     \item We observe `uncertain tokens' in pre-trained MoE models, and we prove that these tokens may select the wrong expert.
     \item We propose a novel fine-tuning method GW-MoE that does not introduce additional inference overhead, helping all experts learn the knowledge of uncertain tokens and reducing the impact of choosing the wrong expert during inference.
     \item GW-MoE outperforms standard fine-tuning across various model scales and natural language tasks.
 \end{itemize}

\section{Background}

\subsection{Mixture of Experts}

Mixture-of-Experts (MoE) is an efficient method for scaling up model sizes~\cite{shazeer2017outrageously, fedus2022switch}. It usually consists of two components: a router $G$ and a set of experts $\{E_1, E_2, ..., E_N\}$. 
In the transformer architecture, MoE is typically employed to replace the feed-forward networks (FFNs) layer. 
Each expert can be regarded as a new FFN, and $G$ determines which $K$ experts to select and computes their respective weights. 
Formally, the output $\boldsymbol{y}$ of an input token $\boldsymbol{x}$ is computed as follows:

\begin{equation}
        \boldsymbol{y} = \sum_{g_i\in TopK(G(\boldsymbol{x}))}g_iE_i(\boldsymbol{x})
        \label{eq:standard}
\end{equation}

\noindent where $g_i$ is the score computed by the router for selecting expert $i$.

Deciding which expert to choose is a difficult discrete optimization problem. 
In addition to the greedy Top-$K$ experts per token shown in Eq.~\ref{eq:standard}, there are many other methods, such as greedy Top-$K$ tokens per expert~\cite{zhou2022mixtureofexperts}, reinforcement learning~\cite{bengio2016conditional}, optimal transport~\cite{liu2023sparsityconstrained}, linear programs~\cite{lewis2021base} and deterministic fixed rules~\cite{roller2021hash, dai2022stablemoe}.

\subsection{Global Workspace Theory}

GWT~\cite{GWT} is an explanation proposed for human cognition. 
It suggests that the human brain has independently functioning modules, and these different modules can compete to send messages into the global workspace. 
Messages in the global workspace can be responded to by other modules; therefore, some complex neural signals can be collaboratively processed by multiple modules.
When learning new knowledge, messages in the global workspace are more readily encoded into long-term memory, enhancing knowledge accessibility.

In MoE, experts are similar to the modules in the human brain, with typically only a small number of experts processing the input tokens. 
However, MoE has no component similar to the global workspace. 
In our method, uncertain tokens can broadcast messages to all experts.

\section{Method}
\label{sec:method}

\textbf{Overview.} 
Inspired by the GWT, we propose GW-MoE. 
The key idea is to broadcast those uncertain tokens to all experts. 
Compared to standard fine-tuning, all experts can learn the knowledge of uncertain tokens, so during inference, uncertain tokens can obtain information from any expert, reducing the impact of incorrect selections.

\subsection{Which Tokens Are Uncertain?}
\label{sec:entropy-uncertain}

The first question to address is what kind of tokens are considered uncertain? 
Intuitively, the tokens that are difficult to determine which $K$ experts to select are uncertain. 
The output of router $G$ can be expressed as follows:

\begin{equation}
    G(\boldsymbol{x}) = [g_0, g_1, ..., g_{N-1}]
    \label{eq:2}
\end{equation}

\noindent $N$ is the total number of experts, and $g_i$ is the score that token $\boldsymbol{x}$ chooses expert $i$. 
The uncertainty of the router $G$ in selecting experts for $\boldsymbol{x}$ can be measured by entropy:

\begin{equation}
    H(\boldsymbol{x}) = -\sum_i g_i log(g_i)
    \label{eq:3}
\end{equation}

Taking $K=1$ as an example, when the router is very certain about choosing expert $j$, there will be $g_j=1, g_{i\neq j}=0$, and the entropy takes the minimum value $0$. 
Conversely, if the router is completely uncertain about which expert to choose, there will be $g_i=\frac{1}{N}, i\in \{0, 1, ..., N-1\}$, and entropy takes the maximum value $logN$.

In Tab~\ref{tab:common-moe}, to compare MoE models with different numbers of experts, we adopt normalized entropy as follows:

\begin{equation}
    H_{norm}(\boldsymbol{x}) = \frac{H(\boldsymbol{x})}{logN}
\end{equation}

\noindent It can be observed that all three models have a portion of tokens that difficult to determine which experts to choose, especially in the JetMoE-8B.

In practice, we measure the entropy distribution of the base model's router output and take the top $5\%$ of the values as the threshold $H^*$.
Tokens with entropy greater than $H^*$ are considered uncertain.

\subsection{GW-MoE}

In standard fine-tuning, the model updates are not fully-differentiable~\cite{puigcerver2024sparse, zhong2024lory}.
Because of the Top-$K$ operator, the gradients of the objective function are only propagated back to the selected experts. 
Therefore, it is difficult to obtain the necessary knowledge when a token cannot choose the correct expert.

GW-MoE enables the expert updates caused by uncertain tokens to be fully-differentiable.
During fine-tuning, the input tokens are divided into certain and uncertain parts based on $H^*$. 
The certain part is processed using the standard MoE approach, as shown in Eq.~\ref{eq:standard}. 
These inputs correspond to simple signals in the human brain, which cannot compete for the right to broadcast themselves to the global workspace. 
The tokens in the uncertain parts will broadcast themselves to all experts as follows:

\begin{equation}
        \boldsymbol{y} = \sum_{i=0}^{N-1}g_iE_i(\boldsymbol{x})
        \label{eq:broadcast}
\end{equation}

Combining Eq.~\ref{eq:standard} and Eq.~\ref{eq:broadcast}, we can get the complete GW-MoE:

\begin{equation}
        \boldsymbol{y} = \left\{\begin{matrix}
             \sum_{i=0}^{N-1}g_iE_i(\boldsymbol{x}), &H(\boldsymbol{x}) \ge H^*\\
             \sum_{g_i\in TopK(G(\boldsymbol{x}))}g_iE_i(\boldsymbol{x}), &H(\boldsymbol{x}) < H^*
        \end{matrix}\right.
        \label{eq:5}
\end{equation}

During inference, all tokens use the standard MoE approach, which means no additional inference overhead is introduced.

\subsection{Implementation Details}

In addition to using $H^*$ to distinguish whether tokens are uncertain, we also introduce an additional hyper-parameter \textit{max num slots}.
In some tasks, we observe that during the initial stages of fine-tuning, the average entropy first increases and then gradually decreases. 
To avoid the increase of training time and memory requirements caused by entropy changes, we use \textit{max num slots} to limit the maximum number of tokens broadcast in each batch. 
We specify the value of \textit{max num slots} based on the average length of the dataset in practice.

\section{Experiments}
\label{sec:exp}

\subsection{Datasets}

We evaluate GW-MoE on multiple datasets across diverse tasks including nature language understanding, question answering, summarization, math problem soving and code generation.
GLUE~\cite{wang2019glue} is a widely used benchmark for testing models’ language understanding capabilities.
It consists of a series of text classification tasks: sentence similarity (STSB;~\citealt{Cer_2017}), (QQP;~\citealt{wang2017bilateral}), (MRPC;~\citealt{dolan2005automatically}), sentiment analysis (SST2;~\citealt{socher2013recursive}), sentence acceptability (CoLA;~\citealt{warstadt2018neural}), natural language inference (MNLI;~\citealt{williams2018broad}), (QNLI;~\citealt{demszky2018transforming}), (RTE;~\citealt{giampiccolo2007third}).
For the summarization task, we use DialogSum~\cite{dialogsum}, which consists of $13460$ dialogues with corresponding manually labeled summaries and topics.
For the question-answering task, we use SQuAD~\cite{squad} and Quoref~\cite{allenai:quoref}.
The former contains questions and answers extracted from Wikipedia articles, designed to assess the machine's ability to understand reading comprehension; the latter is used to test the model's understanding of referential expressions.

In large-scale experiments, we use Alpaca~\cite{alpaca} to instruction-tune the model, which contains a series of human instructions and output pairs. 
We test the model's common sense reasoning on the ARC Challenge~\cite{allenai:arc}, its ability to solve mathematical problems on the GSM8k~\cite{cobbe2021gsm8k}, and evaluate the code generated by the model using HumanEval~\cite{chen2021evaluating}.

\begin{table*}[!ht]
    \vskip  -0.3in

  \centering
  \scalebox{0.98}{
  \begin{tabular}{c|cccccccc|c}
    \toprule
    \textbf{Method}  & \textbf{CoLA} & \textbf{SST-2} & \textbf{MRPC} & \textbf{STS-B} & \textbf{QQP} & \textbf{RTE} & \textbf{QNLI} & \textbf{MNLI} & \textbf{Avg} \\
    \midrule
    Standard FT & $54.65$ & $94.86$ & $\boldsymbol{88.48}$ & $88.62$ & $91.66$ & $67.10$ & $\boldsymbol{92.88}$ & $88.29$ & $83.32$ \\
    GW-MoE & $\boldsymbol{55.17}$ & $\boldsymbol{94.90}$ & $88.35$ & $\boldsymbol{88.96}$ & $\boldsymbol{91.70}$ & $\boldsymbol{70.49}$ & $92.81$ & $\boldsymbol{88.43}$ & $\boldsymbol{83.85}$ \\
    \bottomrule
    
  \end{tabular}
  }
  \caption{Overall comparison on GLUE. For STS-B, we report Pearson Correlattion. For CoLA, we report Matthews correlattion. For others, we report accuracy. The best result on each block is in \textbf{bold}.}
  \label{tab:glue}

  \vskip  -0.1in
\end{table*}

\subsection{Experiments Details}

Firstly, we evaluate GW-MoE on Switch-Base-8~\cite{fedus2022switch}, which is built on T5-Base~\cite{2020t5} with $650$M parameters.
Each MoE layer in Switch-Base-8 contains $8$ experts and activates $1$ experts for each token.
After that, we evaluate our method on the larger-scale model JetMoE-8B~\cite{shen2024jetmoe}.
It also contains $8$ experts in each layer but activates $2$ experts for each token.
Same as~\cite{he-etal-2023-merging}, we fine-tune pretrained base models on selected datasets and report results of the last checkpoint. 
To the best of our knowledge, there is no specialized fine-tuning method proposed for MoE models, and in all experiments, we use standard fine-tuning as our baseline.
Following the recommendations of~\cite{shen2023mixtureofexperts, chi2022representation}, we freeze routers' parameters during fine-tuning.
As discussed in Sec~\ref{sec:entropy-uncertain}, we will select the value of $H^*$  from $1.6$ and $1.8$ for Switch-Base-8 based on the statistical results of the base model on the dataset, and set $H^*$ to $2.0$ in JetMoE.
Other hyperparameters used in our experiments and more details can be found in Appendix~\ref{sec:hyper}.
All experimental results are the average of three runs with different seeds, the standard deviation of the main results is presented in Appendix~\ref{sec:std}.

\subsection{Main Results}
\label{sec:main}

\textbf{GLUE.}
Tab~\ref{tab:glue} compares GW-MoE and the standard fine-tuning on the GLUE benchmark.
Specifically, GW-MoE shows an average $0.53$ increase compared to standard fine-tuning.
The results demonstrate the advantage of GW-MoE in natural language understanding tasks.
It's worth mentioning that we also attempt not to freeze the router's parameters during fine-tuning.
Compared to freezing, it decreases performance in almost all tasks.
This is consistent with~\cite{shen2023mixtureofexperts, chi2022representation}.

\noindent \textbf{NLG Tasks.} Tab~\ref{tab:nlg} shows the comparison results across summarization tasks and question-answering tasks.
Our method achieves better results on all three datasets than standard fine-tuning, with an average improvement of $0.35$.
These results indicate that GW-MoE is effective in NLU tasks and can improve MoE models' performance in natural language generation (NLG) tasks, even in summarization tasks with long inputs.

\begin{table}
    \centering
    \scalebox{0.9}{
    \begin{tabular}{c|ccc}
        \toprule
        \textbf{Method} & \textbf{DialogSum} & \textbf{SQuAD} & \textbf{Quoref} \\
        \midrule
         Standard FT & $24.06$ & $83.35$ & $25.97$ \\
         GW-MoE & $\boldsymbol{24.34}$ & $\boldsymbol{83.52}$ & $\boldsymbol{26.56}$ \\
         \bottomrule
    \end{tabular}
    }
    \caption{Overall comparison on DialogSum, SQuAD and Quoref. For DialogSum, we report Rouge-2 ($\uparrow$). For the question-answering tasks of SQuAD and Quoref, we report the Exact Match ($\uparrow$).}
    \label{tab:nlg}

    \vskip  -0.15in
\end{table}

\subsection{Impact of Additional Computation}

Although GW-MoE does not introduce additional inference costs, broadcasting uncertain tokens during fine-tuning introduces extra computation.
We evaluate the number of samples per second during fine-tuning for different tasks on Switch-Base-8, as shown in Tab~\ref{tab:compu}.
The same task is tested on the same machine, and the batch size is the same.
By setting the \textit{max num slots} based on the average length of the dataset, our method's training speed is only reduced by about $10\%$ compared to standard fine-tuning.
This indicates that our method improves performance and does not introduce significant additional computational costs during training.

\begin{table}
    \centering
    \scalebox{0.9}{
    \begin{tabular}{c|ccc}
        \toprule
        \textbf{Method} & \textbf{SQuAD} & \textbf{Quoref} & \textbf{DialogSum} \\
        \midrule
        Standard FT & $217.7$ & $95.72$ & $119.5$\\
        GW-MoE & $189.4$ & $89.76$ & $110.4$ \\
        \bottomrule
    \end{tabular}
    }
    \caption{The number of samples per second during fine-tuing for different tasks on Switch-Base-8.}
    \label{tab:compu}
\end{table}

\subsection{Performance in Scaling-Size}
\label{sec:IT}

To further verify whether GW-MoE remains effective at a larger scale, we fine-tuned JetMoE-8B on the Alpaca dataset.
Following~\cite{eval-harness}, we test the models' common sense reasoning ability on the Arc-Challenge and its ability to solve mathematical problems on the GSM8K.
Both two tasks are configured with a $5$-shot learning setup.
For code generation, we test models on the humaneval benchmark following~\cite{bigcode-evaluation-harness}.
To fairly compare the code generation capabilities, we uniformly adopt greedy decoding and report pass@$1$.
The results is shown in Tab~\ref{tab:larger-scale}.
In these three tasks, the models fine-tuned by our method all exhibit better performance.
This indicates that GW-MoE can be used not only for models with hundreds of millions of parameters but can also be extended to those with billions of parameters.

\begin{table}
    \centering
    \scalebox{0.9}{
    \begin{tabular}{c|ccc}
        \toprule
        \textbf{Method} & \textbf{Arc} & \textbf{GSM8K} & \textbf{HumanEval} \\
        \midrule
         Standard FT & $48.9$ & $26.1$ & $30.5$\\
         GW-MoE & $\boldsymbol{49.5}$ & $\boldsymbol{27.2}$ & $\boldsymbol{32.3}$ \\
         \bottomrule
    \end{tabular}
    }
    \caption{Comparison of results after fine-tuing on JetMoE-8B. For Arc Challenge, we report accuracy ($\uparrow$). For GSM8K, we report strict match ($\uparrow$). And for HumanEval, we report pass@$1$ ($\uparrow$).}
    \label{tab:larger-scale}

    \vskip  -0.15in
\end{table}

\section{Analysis}
\label{analy}

\subsection{Uncertainty and Wrong Selection.}

In Sec~\ref{sec:intro}, we present our findings: there are some uncertain tokens in the MoE models with billions of parameters, as shown in Tab~\ref{tab:common-moe}.
We hypothesize that \textbf{this uncertainty in the router may lead to the incorrect expert selection.}
To validate our hypothesis, we let the uncertain tokens ($H > 2$) in the final layer of the JetMoE base randomly select experts and set the scores of both selected experts to $0.5$.
We test the model on the three tasks in Sec~\ref{sec:IT}.
One may argue that such repeated tests may lead to test information leaks and an unfair comparison with the baseline. We (1) repeat such experiments x times for each task and report the average performance. 
We also (2) set up control experiments where the same proportion of arbitrary tokens randomly selected experts. These experiments are also repeated with the same seeds in (1).
Such configurations help us separate the effects of lucky improvement from random search.

The results is shown in Fig~\ref{fig:random}.
The three tasks consistently demonstrate the following results: 
1) allowing uncertain tokens to randomly select experts can lead to better performance;
2) allowing the same number of arbitrary tokens to randomly select experts can lead to a decrease in performance.
From this result, we can infer that uncertain tokens may choose the incorrect experts during inference.

\subsection{Global Workspace Broadcasts Knowledge
}
\label{sec:broadcast}

Compared to standard fine-tuning, GW-MoE makes the updates to the experts by uncertain tokens fully-differentiable.
This ensures that during fine-tuning, all experts can learn the knowledge from the uncertain tokens.
In other words, if we consider the experts as memory blocks, we store the information needed by uncertain tokens in all the blocks.
During inference, uncertain tokens that cannot determine the choice of experts can obtain the necessary knowledge from any expert.
Therefore, if GW broadcasts knowledge for uncertain tokens to all experts, we can expect a GW-tuned MoE to perform better than the standard-tuned one when we enforce uncertain tokens to select experts randomly.

To verify this, we conduct experiments of random expert selection on the GW-tuned and standard-tuned Switch-Base-8 models.
To better illustrate the differences, we let all layers' uncertain tokens to randomly select experts.
Tab~\ref{tab:all-experts} shows the results: 
when uncertain tokens in the GW-tuned model randomly select experts, the Exact Match (EM) only decreases $0.78$; 
in contrast, the standard-tuned model suffers a $3.07$ decrease.
This suggests that in the models trained with GW-MoE, \textbf{the knowledge for uncertain tokens is stored in all experts} and they can acquire the necessary knowledge from any expert. Thus, the uncertain tokens are less sensitive to the choice of experts.
Although such broadcasting can lead to redundancies in each expert for the uncertain tokens, we will show it is an effective solution since it's hard to correct the router for the uncertain tokens directly.

\subsection{Router is Hard to Correct}

As GW provides the router with more information about the experts, one may expect GW to bring improvement by helping some uncertain tokens find suitable experts and reduce their choice entropy.
However, we find the ratio of uncertain tokens before and after fine-tuning on Alpaca is approximately $1:3$ on JetMoE when both GW-tuned or standard-tuned.
This indicates it's hard to directly correct the uncertain token problem via tuning even when the router is provided with more information under the current MoE framework.
Therefore, the broadcast knowledge in Sec~\ref{sec:broadcast} brought by GW is the main factor for the improvement.
We leave other methods to solve the uncertain token problem without redundancy for future works.

This also highlights one potential limitation in the existing Top-$K$ routing design: some tokens can't find the best expert combination.
This can also explain why shared-expert~\cite{wu2022residual, dai2024deepseekmoe} design brings benefits: the shared-experts improve the min-max results for tokens that can't find suitable experts.

\begin{table}
    \centering
    \scalebox{0.9}{
    \begin{tabular}{c|cc}
        \toprule
        \textbf{Method} & \textbf{EM} & \textbf{Decrease} \\
        \midrule
         Standard FT & $80.28$ & $3.07$ \\
         GW-MoE & $82.74$ & $0.78$  \\
         \bottomrule
    \end{tabular}
    }
    \caption{The impact of uncertain tokens randomly selecting experts. We report the values of EM on the SQuAD dataset and the decrease compared to using the Top-$K$ selection of experts.}
    \label{tab:all-experts}

    \vskip  -0.15in
\end{table}

\subsection{Uncertain Tokens}

\begin{figure}[htbp]
    \centering
    \includegraphics[width=0.48\textwidth]{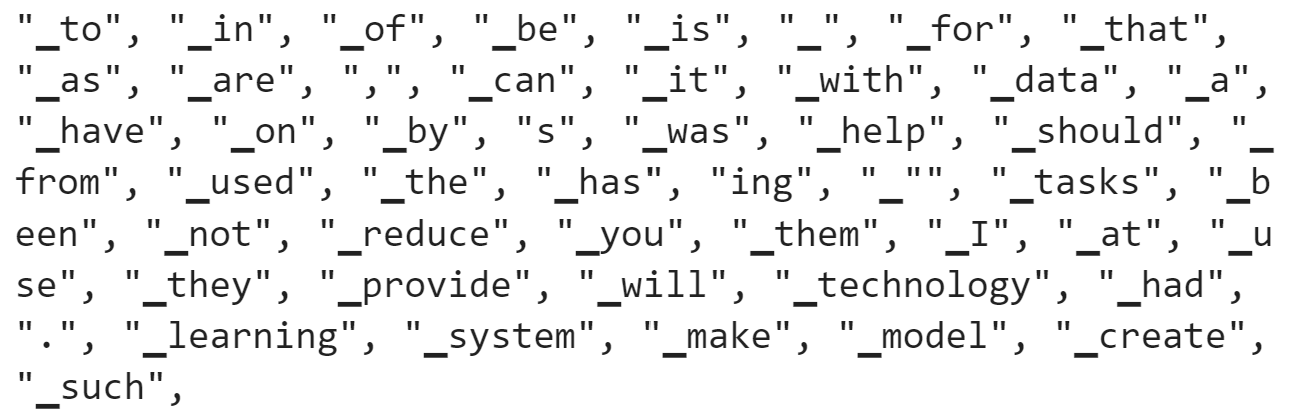}
    \caption{The $50$ most frequently broadcast tokens in JetMoE. Most of them do not have a clear semantic.}
    \label{fig:toks}

    \vskip  -0.15in
\end{figure}

An interesting question is which tokens in natural languages are more likely to be uncertain in MoE models.
We count the $50$ most frequently broadcast tokens in JetMoE on the Alpaca dataset, as shown in Fig~\ref{fig:toks}.
Surprisingly, the most uncertain tokens are those without clear semantics, such as articles, conjunctions, prepositions, punctuation marks, and some very common verbs and nouns.

One possible explanation is that when a model predicts the next token autoregressively, it is indeed difficult to determine which expert to process these words without specific meaning.
For instance, when the word "sing" appears, one might naturally anticipate the next word to be "song." 
However, when the current word is an article like "a", without context, it is impossible to make a prediction because there are too many possible options.
These tokens require different expert knowledge due to different contexts, hence they have higher $H$.

\begin{figure}[htbp]
    \vskip  -0.1in
    \centering
    \includegraphics[width=0.48\textwidth]{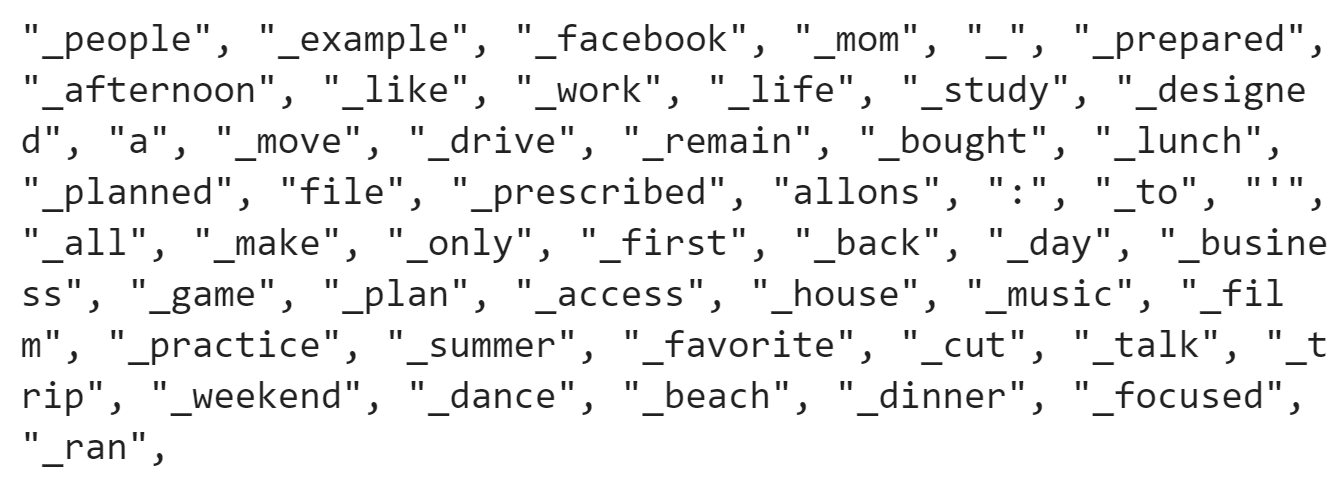}
    \caption{The $50$ most frequently broadcast tokens in the encoder of Switch-Base-8. These tokens are mostly common words with clear semantics.}
    \label{fig:toks-encoder}

    \vskip  -0.15in
\end{figure}

We use the same $H^*$ and also count the most broadcast tokens on the DialogSum dataset by Switch-Base-8.
Since the model is based on the T5 architecture, we perform separate statistics for the encoder and decoder.
The most broadcast tokens in the decoder are highly similar to those in JetMoE, and the most broadcast tokens in the encoder are shown in Fig~\ref{fig:toks-encoder}.
Unlike the decoder, the tokens that are broadcast the most in the encoder include more words with clear semantics.
The role of the encoder is to integrate information, so there is no need to pay special attention to words that lack semantics; instead, common words with clear semantics have higher $H$.

\section{Ablation Study}

\textbf{Activating more experts for all tokens results in worse performance}.
GW-MoE activates more experts for uncertain tokens during fine-tuning. A naive approach would be to activate more experts for all tokens.
We conduct validation experiments on the SQuAD dataset using the Switch-Base-8.
In Tab~\ref{tab:topk+1}, we report the EM of the following four settings: Top1 fine-tuning/Top1 eval, Top2 fine-tuning/Top1 eval, Top1 fine-tuning/Top2 eval and Top2 fine-tuning/Top2 eval.
It can be observed that activating more experts during evaluation than during pre-training can lead to performance drop, even if the number of active experts is changed during fine-tuning.
We believe that not all tokens require the activation of more experts during fine-tuning, only the uncertain tokens need to pass information to all experts.
In addition, it is also important to ensure that the number of activated experts remains consistent with that during pre-training.

\begin{table}
    \centering
    \scalebox{0.9}{
    \begin{tabular}{c|cc}
        \toprule
         & Top1 FT & Top2 FT \\
         \midrule
        Top1 Eval  & $\boldsymbol{83.35}$ & $82.29$ \\
        Top2 Eval & $78.98$ & $81.78$ \\
        \bottomrule
    \end{tabular}
    }
    \caption{EM of different Top-$K$ tuning/inference combinations on SQuAD.
    }
    \label{tab:topk+1}
\end{table}

\noindent \textbf{It's not necessary to broadcast uncertain tokens during inference}.
GW-MoE does not introduce additional inference costs, it remains the same as the standard model during inference.
We test the impact of broadcasting uncertain tokens on performance during inference, as shown in Tab~\ref{tab:inf}.
Interestingly, not broadcasting uncertain tokens is a better choice during inference in almost all tasks.
We speculate that the reason might be the difference in entropy distribution between the training set and the test set, where some tokens that are certain in the training set are incorrectly broadcast during inference.
Unlike~\cite{huang2024harder}, GW-MoE focuses on the uncertainty of expert selection rather than activating more experts for harder tokens.

\begin{table}
    \centering
    \scalebox{0.72}{
    \begin{tabular}{c|cccc}
        \toprule
        & \textbf{GLUE (avg)} & \textbf{SQuAD} & \textbf{Quoref} & \textbf{DialogSum} \\
        \midrule
        w/o braodcast & $\boldsymbol{83.85}$ & $\boldsymbol{83.52}$ & $\boldsymbol{26.56}$ & $\boldsymbol{24.34}$ \\
        w/ braodcast & $83.70$ & $83.48$ & $26.43$ & $24.26$ \\
        \bottomrule
    \end{tabular}
    }
    \caption{Comparison of whether to broadcast uncertain tokens during inference with the same metrics as Sec~\ref{sec:main}.}
    \label{tab:inf}
    \vskip  -0.2in
\end{table}

\begin{figure}[htbp]
    \centering
    \includegraphics[width=0.48\textwidth]{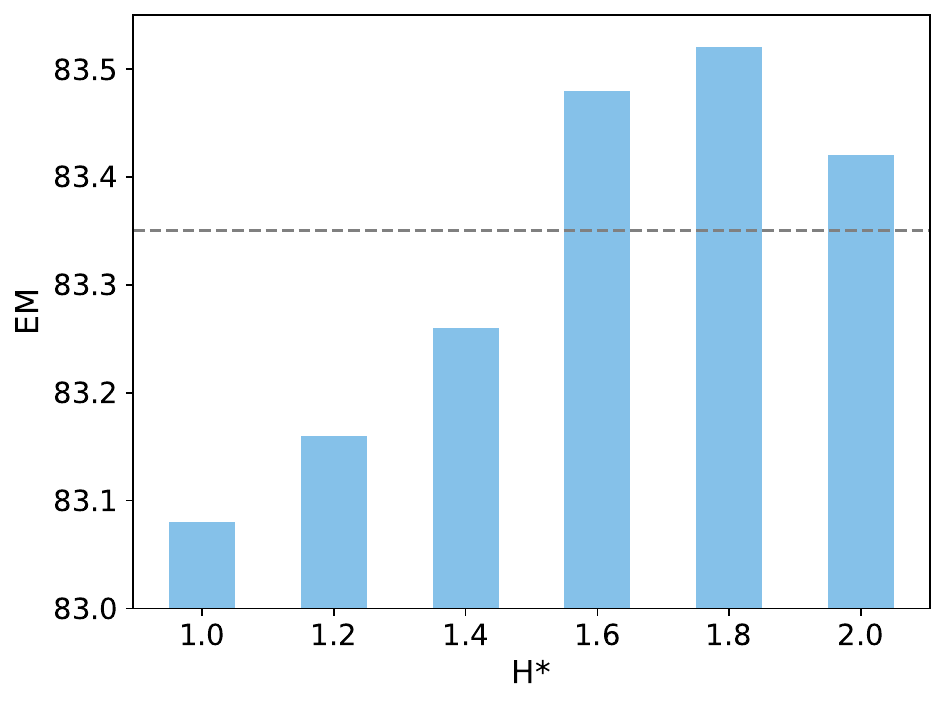}
    \caption{The variation of EM with $H^*$. The dashed line indicates the result of standard fine-tuning.}
    \label{fig:match}

    \vskip  -0.2in
\end{figure}

\noindent \textbf{$H^*$ needs to match the \textit{max num slots}}.
In GW-MoE, two additional hyperparameters are introduced: $H^*$ and \textit{max num slots}. The latter is used to limit the additional computational overhead during fine-tuning.
We fix the encoder's \textit{max num slots} to $16$, decoder's to $1$, and we try different values of $H^*$ on the SQuAD dataset.
The variation of EM with $H^*$ can be seen in Fig~\ref{fig:match}.
EM is highest when $ H^*$ is $1.8$, which is also the value we adopt in our experiments based on the distribution of the encoder's entropy.
When $H^*$ greater than $1.8$, we suspect it could cause the exclusion of uncertain tokens from selection, resulting in a decrease of EM; when $H^*$ is less than $1.8$, due to the limitation of the \textit{max num slots}, certain tokens might occupy the limited broadcasting rights, preventing the truly uncertain tokens from being broadcast.
In summary, the two additional hyperparameters in GW-MoE need to match each other.
We suggest using a value close to $5\%$ of the dataset's average length as the \textit{max num slots}, corresponding to the statistical method of $H^*$ in Sec~\ref{sec:entropy-uncertain}.
In our experiments, such settings typically result in a stable improvement.

\section{Related Work}

\subsection{Mixture of Experts}

\citet{shazeer2017outrageously} introduce the Mixture of Experts(MoE) into the LSTM model and apply it to the machine translation task. Subsequently, \citet{lepikhin2020gshard} is the first to introduce MoE into the transformer model. With the release of Switch Transformer~\cite{fedus2022switch}, MoE begins to be widely used in the training of LLMs, such as ~\cite{jiang2024mixtral, dai2024deepseekmoe, shen2024jetmoe}. At the same time, many works focus on the design of the router: \citet{roller2021hash, dai2022stablemoe} propose using static routing to ensure load balancing among experts and stable training; \citet{zhou2022mixtureofexperts} propose using expert-choice, allowing different tokens to be assigned to various experts.
\citet{liu2023towards, qiu2023emergent} use the experts' first layer weights as expert embedding to further connect the router and expert.
In addition,
~\citet{wu2022residual, rajbhandari2022deepspeedmoe, dai2024deepseekmoe} utilized shared experts to represent the common knowledge among experts.

Recently, many works have suggested that the number of experts should be dynamically determined based on the input tokens~\cite{li2023adaptive, huang2024harder}. 
These works are more relevant to ours; they also select more experts for some tokens based on the router's information.
Unlike they activate more parameters for some tokens during inference, our method focuses on the tokens in the input sequence that are uncertain about the expert choice.
We activate all experts for these tokens only during fine-tuning
while remaining consistent with the standard MoE during inference, thus not introducing additional overhead.

We also find that there is limited research on fine-tuning MoE models. 
\citet{shen2023mixtureofexperts} and~\citet{chi2022representation} suggest that freezing the router parameters can prevent overfitting during fine-tuning. 
\citet{zhao2024hypermoe} introduces an additional hypernetwork to provide information from unselected experts. By learning the parameters of the hypernetwork during fine-tuning, it performs better than the standard MoE.
Our work does not introduce additional parameters and provides a new perspective on routing uncertainty for model fine-tuning.

\subsection{Global Workspace Theory}

GWT~\citep{GWT}, as a theory of consciousness, is receiving growing attention in the quest to build Artificial General Intelligence. 
These works~\citep{vanrullen2021deep, butlin2023consciousness} leverage GWT to discuss how to build true intelligence from existing models. Our work differs from theirs, we focus on how to draw on GWT to make uncertain tokens be able to acquire the required knowledge during inference. 

\section{Conclusion}

In this work, we introduce GW-MoE, a novel MoE model fine-tuning method, which does not introduce any additional inference overhead.
We observe many uncertain tokens in the pre-trained MoE model, and routers can assign worse-than-random experts to them.
Inspired by GWT, we broadcast uncertain tokens during fine-tuning, allowing all experts to learn this part of the knowledge; during inference, the required knowledge can be obtained from any expert.
We show the effectiveness of our method on multiple NLP tasks.
We conduct in-depth analyses of the router behaviors in MoE and prove GWT brings improvement via broadcasting knowledge.
Our analysis can provide insights for the design of routers and MoE pre-training.

\section*{Limitations}

There are several limitations: 1) In this work, we only focus on the models' fine-tuning and did not explore the possibility of using GW-MoE during the pre-training. 2) Due to the limitations of experimental conditions, we did not validate our method on larger-scale models, such as Mixtral $8\times7B$ and Mixtral $8\times 22B$. 3) 
Based on experimental observations and the results from DeepSeek MoE~\cite{dai2024deepseekmoe}, we find that MoE models underperform dense models in understanding tasks like MMLU. Due to the lack of semantics in the broadcast tokens, GW-MoE is also unable to provide an enhancement to decoder-only models in understanding tasks. We will leave the improvements for these issues to future work.

\bibliography{gw}

\appendix

\section{Hyperparameters Used and More Experiments Details}
\label{sec:hyper}

We use Adam optimizer for all tasks with the first $10\%$ warm-up steps.
For GLUE benchmarks, we employ a batch size of $32$, train for $10$ epochs besides RTE ($20$ epochs), and perform a grid search for the appropriate learning rate between $1e-5$ and $5e-5$. 
In our method, the \textit{max num slots} is set to $8$ for encoder, and $1$ for decoder.
In question answering tasks, we adopt a batch size of $64$, a learning rate of $3e-5$, and train for $10$ epochs with \textit{max num slots} allocated as $16$ for the encoder and $1$ for the decoder.
For the summarization task,  we utilize a batch size of $64$, a learning rate of $5e-5$ and train for $10$ epochs.
The \textit{max num slots} is set to $32$ for encoder, and $8$ for decoder.
In large-scale experiments, we use $128$ as batch size, $2e-5$ as learning rate, $16$ as \textit{max num slots} and train $3$ epochs.
For Switch-Base-8, we set the maximum token length to $384$ for question-answering tasks, $256$ for other tasks besides DialogSum.
We set max length to $1024$, and max target lenghth to $256$ for DialogSum.
In larger scale experiments, we set max length to $512$ during instruction-tuning; during evaluation, we set max length to $4096$ and max target length to $512$.

Our all experiments are conducted on eight Nvidia 80GB A100 GPUs, with the running time for different tasks ranging from a few minutes to ten hours.

\renewcommand{\dblfloatpagefraction}{.9}
\begin{table*}[ht]
  \centering
  \scalebox{0.8}{
  \begin{tabular}{c|cccccccc}
    \toprule
    \textbf{Method}  & \textbf{CoLA} & \textbf{SST-2} & \textbf{MRPC} & \textbf{STS-B} & \textbf{QQP} & \textbf{RTE} & \textbf{QNLI} & \textbf{MNLI} \\
    \midrule
    Standard FT & 54.65$\scalebox{0.8}{$\pm$1.92}$ & 94.86$\scalebox{0.8}{$\pm$0.24}$ & 88.48$\scalebox{0.8}{$\pm$0.32}$ & 88.62$\scalebox{0.8}{$\pm$0.21}$ & 91.66$\scalebox{0.8}{$\pm$0.04}$ & 67.10$\scalebox{0.8}{$\pm$0.96}$& 92.88$\scalebox{0.8}{$\pm$0.32}$ & 88.29$\scalebox{0.8}{$\pm$0.03}$  \\
    GW-MoE & 55.17$\scalebox{0.8}{$\pm$1.13}$ & 94.90$\scalebox{0.8}{$\pm$0.28}$ & 88.35$\scalebox{0.8}{$\pm$1.67}$ & 88.96$\scalebox{0.8}{$\pm$0.12}$ & 91.70$\scalebox{0.8}{$\pm$0.09}$ & 70.49$\scalebox{0.8}{$\pm$3.23}$& 92.81$\scalebox{0.8}{$\pm$0.02}$ & 88.43$\scalebox{0.8}{$\pm$0.17}$  \\
    \bottomrule
    
  \end{tabular}
  }
  \caption{The results with standard deviations for the GLUE in section~\ref{sec:main}.}
  \label{tab:glue-std}
\end{table*}

\begin{table}[h]
    \centering
    \scalebox{0.85}{
    \begin{tabular}{c|ccc}
        \toprule
        \textbf{Method} & \textbf{DialogSum} & \textbf{SQuAD} & \textbf{Quoref} \\
        \midrule
         Standard FT & 24.06$\scalebox{0.8}{$\pm$0.28}$ & 83.35$\scalebox{0.8}{$\pm$0.18}$ & 25.97$\scalebox{0.8}{$\pm$0.63}$ \\
         GW-MoE & 24.34$\scalebox{0.8}{$\pm$0.33}$ & 83.52$\scalebox{0.8}{$\pm$0.15}$ & 26.56$\scalebox{0.8}{$\pm$0.07}$ \\
         \bottomrule
    \end{tabular}
    }
    \caption{The results with standard deviations for NLG tasks in section~\ref{sec:main}.}
    \label{tab:nlg-std}
\end{table}

\section{Standard Deviation of the Main Results}
\label{sec:std}

Tab~\ref{tab:glue-std} and Tab~\ref{tab:nlg-std} demonstrate the standard deviations of the results for GLUE and the NLG tasks in section~\ref{sec:main}.

\end{document}